\documentclass[letterpaper, 10 pt, conference]{ieeeconf}  

\IEEEoverridecommandlockouts                              

\usepackage{amssymb}
\usepackage{graphicx}
\usepackage{float}  
\usepackage{placeins}
\usepackage{amsmath}
\usepackage{subcaption}
\usepackage{cite} 
\usepackage{mathtools}
\usepackage[hidelinks]{hyperref}
\usepackage{booktabs}
\usepackage{xcolor}
\usepackage{tikz}
\usetikzlibrary{shapes.geometric, arrows.meta, shapes.arrows, positioning, fit, calc}

\usepackage[font=small]{caption}   
\title{\textbf{GUARD: Sensitivity-aware Uncertainty Attribution \\for Safe Traditional and Learning-based Robot Control}}

\author{Woo-Jeong Baek$^{1,*}$, Johannes A. Gaus$^{2,*}$, Jonas Kiemel$^{3}$, Junheon Yoon$^{5}$,\\
Suhan Park$^{4}$, Tamim Asfour$^{3}$, and Jaeheung Park$^{5}$%
\thanks{$^{*}$Equal contribution.}%
\thanks{$^{1}$ Artificial Intelligence Institute (AIIS), Seoul National University, Republic of Korea.}
\thanks{$^{2}$ Hertie Institute for Clinical Brain Research \& Center for Integrative Neuroscience \& Center for Bionic Intelligence Tübingen Stuttgart (BITS), University of Tübingen, Germany.}
\thanks{$^{3}$ High Performance Humanoid Technologies Lab, Institute for Anthropomatics and Robotics (IAR), Karlsruhe Institute of Technology (KIT), Germany.}%
\thanks{$^{4}$ School of Robotics, Kwangwoon University, Seoul, Republic of Korea.}%
\thanks{$^{5}$ Department of Intelligence and Information, Graduate School of Convergence Science and Technology, Seoul National University, Republic of Korea.}%
\thanks{The authors thank the International Max Planck Research School for Intelligent Systems (IMPRS-IS) for supporting Johannes A. Gaus.}
}

\begin{document}

\maketitle
\thispagestyle{empty}
\pagestyle{empty}

\begin{abstract}
This paper presents the framework \textbf{GUARD} (\textbf{G}uided robot control via \textbf{U}ncertainty attribution and prob\textbf{A}bilistic kernel optimization for \textbf{R}isk-aware \textbf{D}ecision making) that combines a technique from probabilistic state estimation with an uncertainty identification method to derive critical thresholds for robot control algorithms. The first part focuses on complementing a method for probabilistic kernel optimization (PKO) by a sensitivity-aware uncertainty attribution technique. Next, the estimated uncertainties and their sources are forwarded to the robot controllers to adapt safety margins and trigger the initialization of safe backup trajectories. To validate GUARD's applicability and performance, experiments are conducted on a traditional model predictive control (MPC) algorithm and a reinforcement learning (RL) based controller. The results demonstrate that GUARD reduces collision rates by 44--67\% in the RL-based evaluation, while decreasing backup policy activation by 50--75\% in the medium- and high-uncertainty profiles. In the MPC simulation, GUARD successfully avoids collisions. By unifying the probabilistic kernel optimization algorithm from robot state estimation with the uncertainty estimation method for learning-based control, GUARD brings together traditional techniques and learning for safe and efficient robot applications. 
\end{abstract}

\section{Introduction} \label{sec:intro}
\noindent The growing trend toward flexible real-world robot systems requires methods that ensure robustness against undesired disturbances while assuring safety during operation. Here, the term \emph{robustness} stands for the system's ability to maintain its performance and functionality despite the occurrence of undesired disturbances or dynamical variations. In contrast, \emph{safety} at system runtime refers to the robot's ability to adapt its motion online to keep the occurrence probability for risks below a specified threshold. In this manuscript, we focus on collisions as risk events.
While a long line of works for collision avoidance exists in robotics literature, the applicability of the majority of these is limited to the respective subfields. For example, collision avoidance methods derived for robot navigation mostly neglect how the inaccuracies from the state estimation affect the collision probability. Similarly, end-to-end learning control frameworks or learned belief spaces tend to underlie assumptions on the training data. However, with the increasing desire to develop complex robot applications that employ methods from various subareas, studies on combining different techniques in safe and efficient manner are becoming increasingly important. 
\begin{figure}[t!]
    \centering
    \includegraphics[width=1.\linewidth]{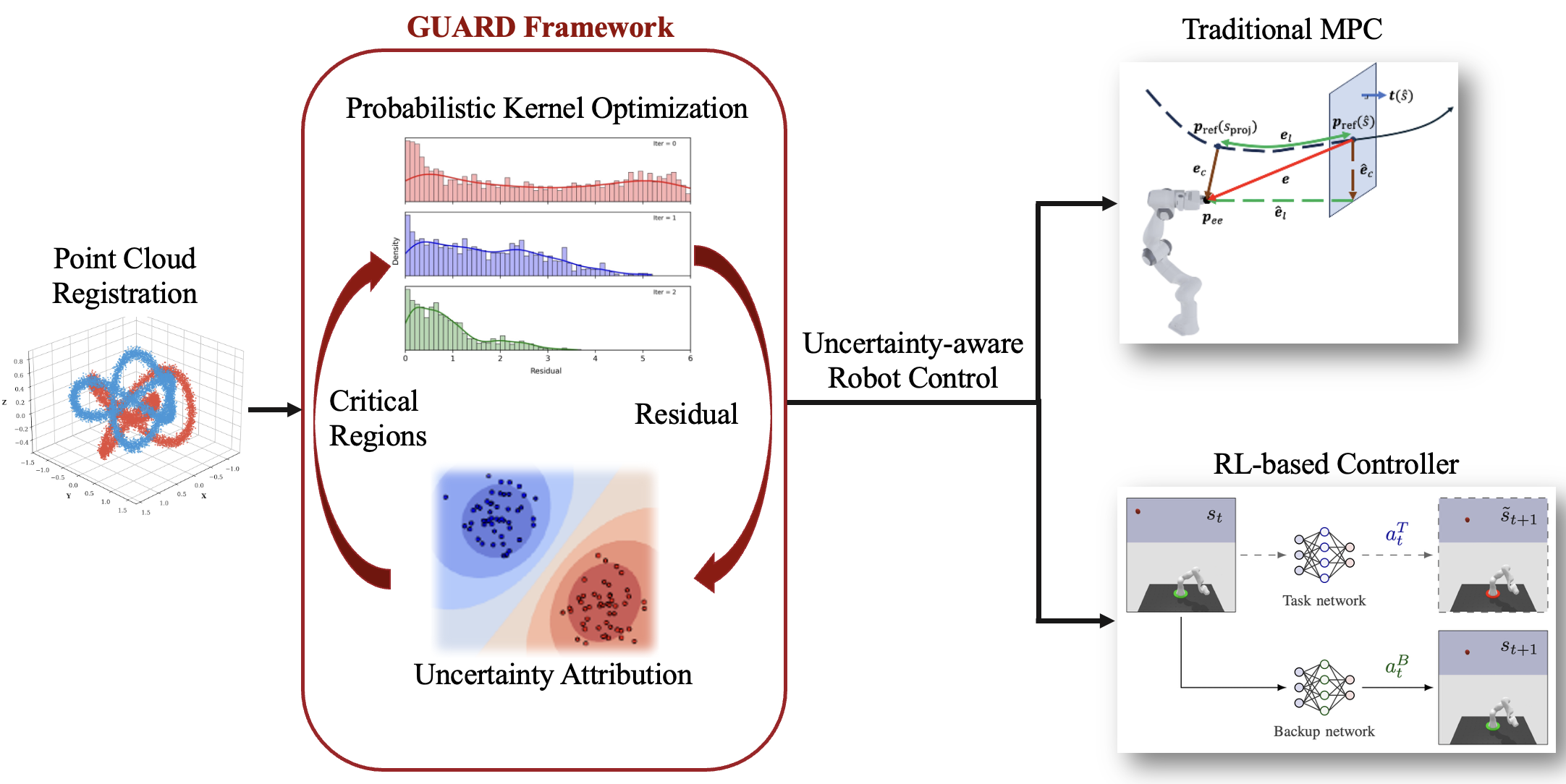}
	\caption{GUARD combines a kernel optimization technique with an uncertainty attribution method to develop safe robot control frameworks. The validation experiments on a MPC algorithm and a reinforcement learning (RL)-based controller show that GUARD successfully avoids collisions while preserving efficiency. Figures adapted from \cite{Choi2025}, \cite{Yoon2025}, \cite{Kiemel2024SafeMotionsRisk}.}  
	\label{fig:Fig1}
\end{figure}
Motivated by this, we present the framework GUARD (\textbf{G}uided robot control via \textbf{U}ncertainty attribution and prob\textbf{A}bilistic kernel optimization for \textbf{R}isk-aware \textbf{D}ecision making) that determines the occurrence probability for collisions by determining the uncertainties from the robot perception and state estimation. Briefly, the probabilistic kernel optimization (PKO) method of Choi et al. in \cite{Choi2025} proposes to enhance the efficiency in the robot state estimation by assigning weights to residuals due to outliers. Here, the residual reflects the discrepancy between the modeled state and the actual state obtained from the measurement data in real-time. By interpreting the residual of the PKO algorithm as the uncertainty, GUARD complements PKO with the technique TCAV-GPC in \cite{gaus2025humaninterpretableuncertaintyexplanationspoint}, which identifies root causes for the uncertainties. As shown in Fig. \ref{fig:Fig1}, this combination enables the PKO to focus on regions with critical uncertainty sources, thereby allowing the identification of regions with high risks.
Next, the uncertainty with its root cause is integrated into two robot control frameworks to avoid collisions in real-time. Here, the uncertainty sources are identified via the so-called concept activation vectors introduced in \cite{Kim2018}. Specifically, these vectors are derived on the basis of the sensitivity, that captures how much an algorithm's output varies due to  disturbances applied on the respective input data. Formally, the partial derivative is one possibility to express the sensitivity. 
\begin{figure*}[t!]
    \centering
    \definecolor{box_blue}{HTML}{98BFEF}
\definecolor{box_yellow}{HTML}{FBE7A3}
\definecolor{box_green}{HTML}{CADFB8}
\definecolor{arrow_red}{HTML}{B02418}
\hspace*{-0.2cm}
\begin{tikzpicture}[scale=0.48, transform shape,
    >=Latex,
    node distance=1.8cm and 2.2cm,
    every node/.style={font=\small},
    blueoval/.style={ellipse, draw=none, fill=box_blue, align=center, minimum width=3.2cm, minimum height=1.4cm},
    yellowoval/.style={ellipse, draw=none, fill=box_yellow, align=center, minimum width=3.2cm, minimum height=1.4cm},
    greenoval/.style={ellipse, draw=none, fill=box_green, align=center, minimum width=3.2cm, minimum height=1.4cm},
    bluebox/.style={rectangle, rounded corners=3pt, draw=none, fill=box_blue, align=center, minimum width=2.7cm, minimum height=1.2cm},
    yellowbox/.style={rectangle, rounded corners=3pt, draw=none, fill=box_yellow, align=center, minimum width=2.7cm, minimum height=1.2cm},
    greenbox/.style={rectangle, rounded corners=3pt, draw=none, fill=box_green, align=center, minimum width=2.7cm, minimum height=1.2cm}, 
    central/.style={
        rounded corners=15pt,
        draw=red!70!black,
        thick,
        outer sep=0pt,
        minimum width=7cm,
        minimum height=4.5cm,
        align=center
    }]

\tikzset{
    connection/.style={
        draw=black,
        semithick,
    }
}

\tikzset{
    arrow_connection/.style={
        draw=black,
        semithick,
        -{Stealth[length=1mm 0.5]},
    }
}

\node[bluebox] (state) {State\\Estimation};
\node[yellowbox, below=0.8cm of state] (perception) {Perception};

\node[blueoval, above left=of state, xshift=0.2cm] (photometrics) {Photometrics,\\Prob. Methods\\(ROVIO, EKF, ..)};
\node[blueoval, above=of state, xshift=-1cm] (offline) {Offline\\(GTSAM, \\iSAM2, ..)};
\node[blueoval, left=of state, yshift=1cm, xshift=0.1cm] (dnn) {DNN\\Methods \cite{Yang2020}, \cite{Wang2017}};

\draw[connection] (photometrics) -- (state);
\draw[connection] (offline) -- (state);
\draw[connection] (dnn.0) -- (state.180);

\node[yellowoval, left=of perception, yshift=-0.8cm, xshift=0.2cm] (registration) {Global, Local\\Registration\\\cite{Zhou2016FGR},\cite{Yang2021TEASER}};
\node[yellowoval, below=of perception, xshift=-0.5cm, yshift=0.3cm] (prob) {Probabilistic\\Methods \cite{gaus2025humaninterpretableuncertaintyexplanationspoint}, \cite{Maken2021}};
\node[yellowoval, below left=of perception] (learning) {Learning\\Methods \cite{Landry2019}};

\draw[connection] (registration) -- (perception.180);
\draw[connection] (prob) -- (perception);
\draw[connection, shorten >=-1pt] (learning.25) -- (perception.204);

\node[scale=0.19, xshift=7.0cm,
  single arrow,          
  draw=arrow_red, fill=white, 
  minimum width=7cm, minimum height=0.1cm, 
  single arrow tip angle=100, 
  inner sep=-10pt,
  single arrow head extend=0.4cm, 
  text width=9cm, align=center 
] at ($(state.east)!0.5!(perception.east)$) (red_arrow)  {\fontsize{50}{50}\selectfont  \textbf{GUARD}};

\node[greenbox, right=0cm of red_arrow] (risk) {Risk-aware\\Control};
\draw[connection] (state.east) -- (red_arrow.160);
\draw[connection] (perception.east) -- (red_arrow.200);

\node[greenoval, above=0.4cm of risk, xshift=1.8cm] (traditional) {Traditional\\Control \cite{Ames2019}, \cite{Singletary2021}};
\node[greenoval, above=0.5cm of traditional, xshift=1cm] (mpc) {Path or Time\\Optimal MPC\\ \cite{Romero2022}, \cite{Yoon2025}};
\node[greenoval, right=0.5cm of traditional, yshift=0.6cm] (sampling) {Sampling-based\\Methods};
\node[greenoval, below=0.5cm of risk, xshift=1.0cm] (learningc) {Safe Reinforcement\\Learning};
\node[greenoval, below=0.5cm of learningc, xshift=1cm] (e2e) {Model-based\\Methods \cite{fisac2018general, wabersich2021predictive}};
\node[greenoval, right=0.55cm of learningc, yshift=-0.6cm] (rl) {Model-free\\Methods \\\cite{Kiemel2024SafeMotionsRisk, thananjeyan2021recovery}};

\draw[connection] (risk) -- (traditional);
\draw[connection] (risk.270) -- (learningc.130);
\draw[connection] (traditional) -- (mpc);
\draw[connection] (traditional.0) -- (sampling);
\draw[connection] (learningc.270) -- (e2e);
\draw[connection] (learningc.0) -- (rl);

\node[] (state-of-the-art) at (3.3cm, 5.3cm) {\fontsize{18}{18}\selectfont \textbf{State-of-the-Art Literature}};
\node[right=10.9cm of state-of-the-art, yshift=-0.5cm, align=center] (contribution) {{\fontsize{18}{22}\selectfont \textbf{Our Contribution:}} \\[1.0em] {\fontsize{18}{22}\selectfont \textbf{The GUARD Framework}}};

\draw[dashed, thick] (12.9cm, 3.6cm) -- +(0, -9cm);

\node[central] (guard) at (21.5cm, -1cm) {};

\node[blueoval, above=0.4cm of guard, xshift=-6.5cm] (pse) {Probabilistic\\State\\Estimation \cite{Choi2025}};
\node[yellowoval, below=0.4cm of guard, xshift=-6.5cm] (uap) {Uncertainty-\\aware\\Perception \cite{gaus2025humaninterpretableuncertaintyexplanationspoint}};

\node[greenoval, above=0.4cm of guard, xshift=7cm] (mpc) {MPC-based \\Controller from\\Yoon et al.\cite{Yoon2025}};
\node[greenoval, below=0.4cm of guard, xshift=7cm] (rl) {RL-based \\Controller from\\Kiemel et al.\cite{Kiemel2024SafeMotionsRisk}};

\node[above=1.45cm of guard.center, font=\bfseries\color{arrow_red}, scale=1.1] 
    {GUARD Framework};

\node[align=center, scale=1.2] at ([xshift=-1.45cm,yshift=0.6cm]guard.center) 
    {Residuals\\(State\\Estimation)};

\node[align=center, scale=1.2] at ([xshift=-1.5cm,yshift=-1.3cm]guard.center) 
    {Uncertainty\\Attribution};

\node[align=center, scale=1.2] at ([xshift=2.1cm,yshift=0.2cm]guard.center) 
    {CBF\\Formulation};
\node[align=center, scale=1.2] at ([xshift=2.1cm,yshift=-0.9cm]guard.center) 
    {Risk\\Threshold};
    
\draw[-{Stealth[length=1.5mm 0.5]}, thick, red!40!black] 
    ([xshift=-2.67cm,yshift=-1.3cm]guard.center)
    arc[start angle=220,end angle=130,radius=1.32cm];
\draw[-{Stealth[length=1.5mm 0.5]}, thick, red!40!black] 
    ([xshift=-0.35cm,yshift=0.5cm]guard.center)
    arc[start angle=40,end angle=-50,radius=1.32cm];
\draw[arrow_connection, black, thick] ([xshift=0.3cm,yshift=0cm]guard.center) -- + (0.6cm, 0cm);
\draw[arrow_connection, black, thick] ([xshift=0.3cm,yshift=-1.0cm]guard.center) -- + (0.6cm, 0cm);


\draw[arrow_connection, shorten >=-1pt] (pse) -- (guard);
\draw[arrow_connection, shorten >=-1pt] (uap) -- (guard);
\draw[arrow_connection] (guard) -- (mpc);
\draw[arrow_connection] (guard) -- (rl);

\end{tikzpicture}
    \vspace{0.02cm}
	\caption{Overview of State-of-the-Art literature and our contribution. While several collision avoidance methods have been presented in the robotic subfields, research on effectively combining methods from state estimation, control and perception remains sparse. GUARD provides one step toward the unification of methods by bringing together a probabilistic kernel optimization algorithm with an uncertainty attribution technique. The uncertainty with its source is fed into control frameworks to exploit the efficiency of learning while ensuring safety via explainable methods.} 
	\label{fig:Fig2}
\end{figure*}
The first validation experiment is performed on the path-following MPC framework in \cite{Yoon2025}, where the uncertainty provided by GUARD acts as a threshold in the Control Barrier Function (CBF). Second, the output of the PKO-ICP module is implemented in a reinforcement learning (RL)-based controller in \cite{Kiemel2024SafeMotionsRisk}. Once a specified threshold of the uncertainty is met, the controller switches to a learned backup policy that generates a collision-free trajectory. The experimental findings show that GUARD improves safety by reducing collision rates by 44--67\% compared with magnitude-only uncertainty handling. Furthermore, GUARD reduces unnecessary backup policy activations by 50--75\% in selected uncertainty profiles. 
The remainder of this paper is structured as follows: After summarizing related works in the field of state estimation, perception and control in Section \ref{sec:sota}, Section \ref{sec:preliminaries} introduces the PKO algorithm, TCAV-GPC and the control algorithms. Afterward, Section \ref{sec:method} presents the framework GUARD. Section \ref{sec:experiments} explains the experiments on the two control frameworks and discusses the results. Section \ref{sec:conclusion} concludes the main findings of this paper.

\section{Related Work} \label{sec:sota}
\subsection{State Estimation}
\noindent In robotics literature, state estimation problems are commonly formulated as inference problems with residuals between the model and the measurement data of the robot state. To solve these problems, existing works suggest Kalman Filter based approaches like recursive filtering with covariance propagation in \cite{Mourikis2007} and nonlinear least-squares smoothing or MAP estimation over keyframes. Furthermore, Leutenegger et al. present a sliding window method in \cite{Leutenegger2015,Qin2018}. 
In Visual Inertial Odometry (VIO) and SLAM, Forster et al. in \cite{Forster2017} show that the IMU preintegration provides efficient inertial residuals on manifolds. On the other hand, incremental smoothing methods like iSAM2 enable real-time updates via factorizations in  \cite{Kaess2012iSAM2}. Furthermore, the use of deep neural networks (DNNs) have been proposed in \cite{Wang2017}, \cite{Yang2020}. However, to maintain safe robot behavior, collision avoidance point estimates might be insufficient. In the context of safety, methods that enable to monitor the accuracy and the estimator consistency are desired. Here, the analyses performed by Huang in \cite{Huang2010} highlight that linearization and unobservable modes can yield overconfident covariances in EKF-SLAM/VIO. This emphasizes the necessity for refined approaches for safe robot control. 
Suenderhauf et al. in \cite{Suenderhauf2012} introduce a method that is robust against outliers on the basis of switchable constraints. In contrast, more recent approaches like the paper of Choi et al. in \cite{Choi2025_2} argue for the importance of runtime reliability by adapting weights from measurement data. This work is further extended in \cite{gaus2025superframeworksensitivitybased} by deriving risk indicators from backend variables.

\subsection{Perception Algorithms}
\noindent ICP remains the workhorse for point-cloud registration \cite{Besl1992,Chen1992}. Registration uncertainties have been studied via closed-form covariances \cite{Censi2007} and their limitations \cite{Bonnabel2016}, learned covariance predictors (CELLO-3D) \cite{Landry2019}, and probabilistic/multi-modal variants (e.g., Stein-ICP) \cite{Maken2021}. Beyond local ICP, robust and global methods address partial overlap and outliers via global objectives and certifiable estimators (e.g., Fast Global Registration and TEASER/TEASER++) \cite{Zhou2016FGR,Yang2021TEASER}, while probabilistic registration can be formulated through density alignment as in Coherent Point Drift \cite{Myronenko2010CPD}. GUARD aligns perception uncertainty with robust estimation and IRLS interpretations of kernelized losses (e.g., Black--Rangarajan duality) \cite{Black1996Unification}, and connects perception-driven uncertainty to safety margins via stochastic CBF formulations \cite{Clark2021StochasticCBF}.
Most existing approaches quantify uncertainty magnitude but do not analyze its source. A post-hoc ICP attribution method was recently presented in \cite{Qin2024}, but its computational cost limits real-time control integration. GUARD therefore builds on \cite{gaus2025humaninterpretableuncertaintyexplanationspoint}, which uses TCAV \cite{Kim2018} with point-cloud encoders (PointNet/DGCNN) \cite{Qi2017,Wang2019} to assign uncertainty sources; combined with PKO, this yields real-time capability.
The PKO algorithm developed in \cite{Choi2025} adapts robust kernels by matching model/data inlier distributions to improve registration robustness.

\subsection{Model-based Control}
\noindent Classical path following can be split into time--parameterized planners/trackers and path--parameterized MPC. The former optimizes a time-indexed trajectory that a low--level controller tracks \cite{Elbanhawi2014,Karaman2011,Schulman2014,Rakita2018,Nicolis2020,Wang2024}; the latter optimizes progress along a spatial path to gain temporal flexibility and real--time reactivity, evaluated on vehicles, UAVs, and manipulators \cite{Romero2022,Duijkeren2016}. For safety, artificial potential fields have been widely used, while CBFs yield forward-invariance guarantees and typically less intrusive avoidance \cite{Ames2019,Singletary2021}. RMPCC combines path--parameterized MPC with CBFs and achieves high servo rates via Jacobian linearization and Gauss--Newton/SQP \cite{Yoon2025}. 

\subsection{Learning-based Control}
\noindent In reinforcement learning (RL), safety can be enforced by safety filters adjusting unsafe actions generated by an RL policy.  
One line of research draws safety filters from model-based control methods, including Hamilton–Jacobi reachability  analysis~\cite{fisac2018general}, control barrier functions (CBFs)~\cite{cheng2019end}, and model predictive control (MPC)~\cite{wabersich2021predictive}.
Another line of research employs predefined or learned backup policies to override unsafe actions from the RL policy (~\cite{kiemel2022learning, thananjeyan2021recovery, yang2022safe, Kiemel2024SafeMotionsRisk}). The decision to engage the backup policy can be informed by background simulations~\cite{kiemel2022learning}, model knowledge~\cite{yang2022safe}, or risk estimators trained on prior safety violations (~\cite{thananjeyan2021recovery, Kiemel2024SafeMotionsRisk}). 
For scenarios with moving obstacles, the latter are particularly effective, as they naturally handle non-deterministic obstacle behavior. 
\\
\\
\noindent \textbf{Scientific Contribution} \\
\noindent To date, research on combining methods from different robotic subdomains to develop safe robot systems remains sparse. GUARD addresses this research gap by 
\begin{enumerate}
    \item  unifying a probabilistic kernel optimization method from robot state estimation with a perception uncertainty attribution technique 
    \item and transferring the uncertainty with its root cause to robot controllers to adapt safety margins online while preserving the task success rates. 
\end{enumerate}
Practically, the unification of the uncertainty attribution technique with the probabilistic kernel optimization acts as a guide for the controllers. Specifically, GUARD formulates the uncertainty as threshold for the MPC controller in \cite{Yoon2025} and the collision probability in the SafeMotionsRisk framework in \cite{Kiemel2024SafeMotionsRisk}. In a broader sense, GUARD bridges different robotic subdomains as shown on the right part of Fig. \ref{fig:Fig2}.

\begin{figure}[t!!]
\centering
\includegraphics[width=0.45\textwidth]{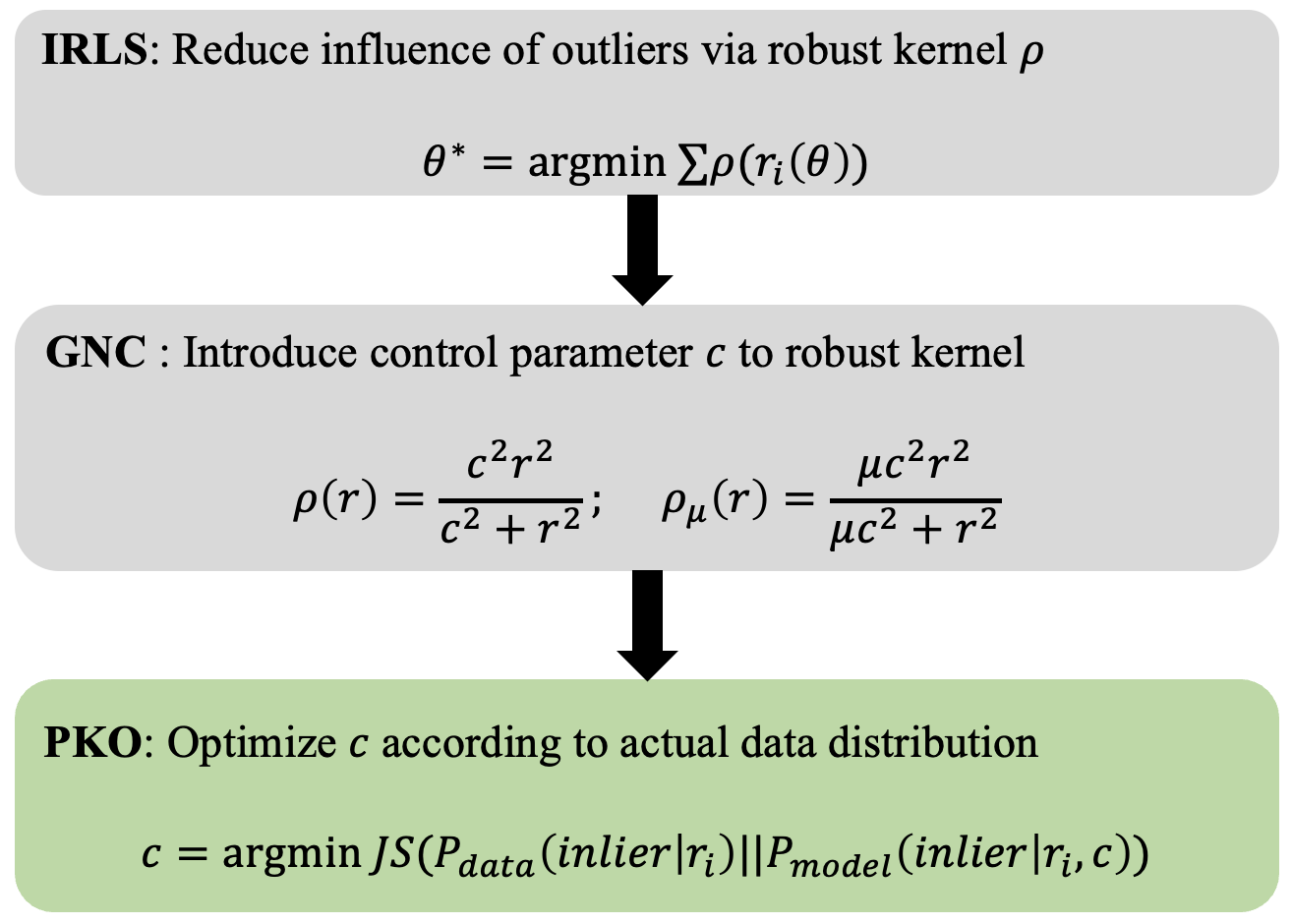}
\caption{Schematic description of the PKO in \cite{Choi2025}. PKO assigns weights to solve non-convex optimization problems.}
\label{fig:pko}
\end{figure}

\section{Preliminaries} \label{sec:preliminaries}

\noindent \textbf{Iterative Closest Point (ICP) Algorithm}\\
ICP estimates the rigid motion $T=(R,t)\in SE(3)$ that aligns a source point cloud to a target cloud by alternating correspondence assignment and a small weighted least-squares update. 
In the point-to-plane variant \cite{Besl1992}, each iteration minimizes residuals of the form 
$$r_i=\mathbf n_i^\top\!\big(R\mathbf p_i+t-\mathbf q_i\big).$$ 
The per-iteration residual set $\{r_i\}$, inlier counts, and overlap proxies provide both the pose estimate and rich cues about the scene (e.g., occlusion, poor overlap, dynamic objects).


\begin{figure} [b!!]
\centering
  \includegraphics[width=\linewidth]{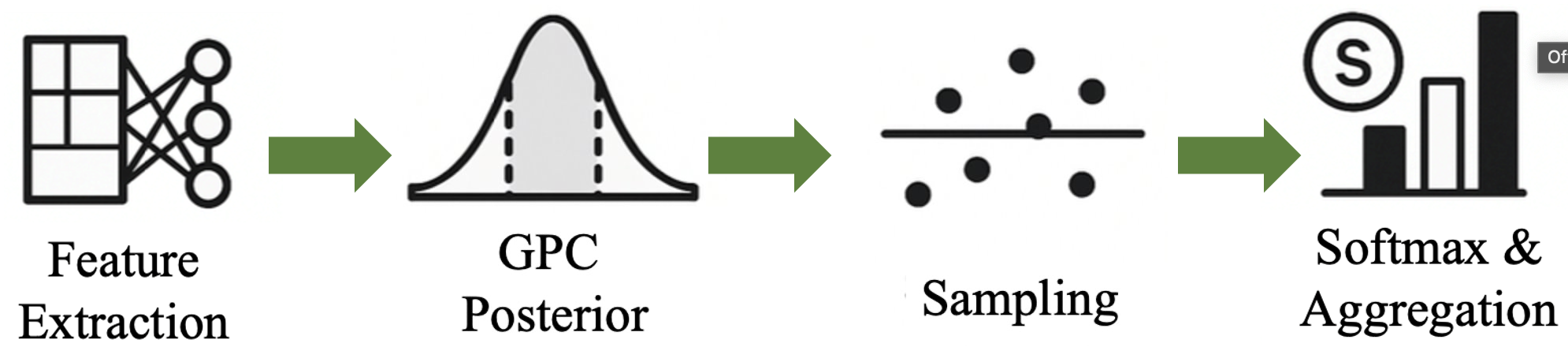}
  \caption{Pipeline of the TCAV-GPC module: The TCAV-GPC algorithm provides uncertainty attributions via concept scores derived on the basis of the technique in \cite{Kim2018}. }
  \label{fig:TCAVGPC}
\end{figure}

\noindent\textbf{Probabilistic Kernel Optimization (PKO)}\\
The framework \cite{Choi2025} refers to the above residuals and suggests a probabilistic weighting scheme for the robust kernel parameters. To be specific, PKO employs the iteratively reweighted least squares (IRLS) algorithm that aims to reduce the influence of infrequent outliers on the optimization process. A robust kernel is introduced according to the Graduated Non-Convexity (GNC) technique. Choi et al. derive the scaling parameter $c$ of the robust kernel on the basis of the actual data distribution. Particularly, a model-based inlier probability $P_{\mathrm{model}}(\text{inlier}\mid r,c)=w(r;c)/Z_w$, and a data-based inlier probability $P_{\mathrm{data}}(\text{inlier}\mid r)$ from a short window of recent residual magnitudes are taken as the basis for the derivation of $c$
At each IRLS iteration, $c$ is adapted via the Jensen-Shannon divergence $JS$:
\begin{equation}
\label{eq:PKO}
c^\star=\arg\min_{c}\ \operatorname{JS}\!\Big(P_{\mathrm{data}}(\cdot)\ \|\ P_{\mathrm{model}}(\cdot\mid c)\Big). 
\end{equation}
This aligns the effective inlier boundary with the data distribution obtained from the actual state. Obviously, the value of $c^\star$ decreases with an improving alignment. PKO improves the tolerance against initialization uncertainties, sensor noise and partial overlaps. \\ 
\\
\noindent\textbf{TCAV–GPC for Uncertainty Attribution}\\
The lack of interpretability in residual-driven adaptation is addressed by integrating the technique of Testing with Concept Activation Vectors (TCAV) \cite{Kim2018}. This TCAV--GP concept-attribution pipeline is taken from the method for human-interpretable uncertainty explanations for point cloud registration \cite{gaus2025humaninterpretableuncertaintyexplanationspoint}. The perception is performed via RGB-D frames that are processed with ICP. With the obtained point clouds, TCAV extracts expressive features via a pretrained model like Dynamic Graph CNN (DGCNN) or a PointNet approach \cite{Qi2017, Wang2019}. 
For the embedding $\boldsymbol e_k=f_\theta(P_k)$ generated by the encoder corresponding to the point cloud $P_k$, a feature vector
\begin{equation}
\boldsymbol\phi_k=\big[\boldsymbol e_k\,]\in\mathbb{R}^d
\end{equation}
is formed. By training a multi-class Gaussian Process Classifier (GPC) with these vectors on artificially perturbed data, the goal is to classify $\boldsymbol\phi_k$ in three semantic uncertainty sources 
$\mathcal C=\{\text{sensor noise},\,\text{pose error},\,\text{partial overlap}\}$ that reflect common origins for uncertainties in practical robot applications. 
At test time the GPC yields calibrated posteriors
\begin{equation}
\pi_k(c)\;=\;p\!\left(c\,\middle|\,\boldsymbol\phi_k,\mathcal D\right);\;\; 
u_k\;=\;-\sum_{c\in\mathcal C}\pi_k(c)\log \pi_k(c),
\end{equation}
where $u_k$ summarizes the predictive uncertainty and $c\in\mathcal C$. 
In practice, $\pi_k(c)$ is computed via a standard multi-class GP link with a Laplace approximation or Monte Carlo samples of the latent scores. Sampling from the GPC posterior provides both per-concept attribution scores $\{\pi_k(c)\}_{c\in\mathcal C}$ and calibrated uncertainty estimates $u_k$. 
This analysis, inspired by TCAV in \cite{Kim2018} in the encoder space, provides 
$
s_c(\boldsymbol e_k)\;=\;\nabla_{\boldsymbol e}\,g_c(\boldsymbol e)\big|_{\boldsymbol e=\boldsymbol e_k}^{\!\top}\mathbf v_c,
$
where $g_c$ is the pre-link class score and $\mathbf v_c$ is the concept activation vector for concept $c$. 
This enables the system both to measure \emph{how} uncertain a registration is ($u_k$) and to explain \emph{why} (via the dominant concept $c_k^\star=\arg\max_{c}\pi_k(c)$). In GUARD, these act as thresholds for the PKO.\\

\noindent\textbf{Risk-aware Control Frameworks}\\
We evaluate GUARD with the following two controllers.\\

\begin{figure}[b]
\centering
\includegraphics[width=0.35\textwidth]{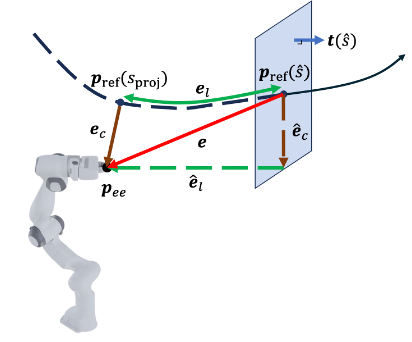}
\caption{The RMPCC controller parametrizes the robot path via a path parameter $s$. The error is split into a contouring and a lag error by referring to the local path tangent. Fig. taken from \cite{Yoon2025}. }
\label{fig:rmpcc}
\end{figure}

\noindent \textbf{Reactive Model Predictive Contouring Control (RMPCC)}\\
Yoon et al. in \cite{Yoon2025} present the RMPCC 
that follows a reference path with CBFs. In particular, the robot path is parameterized by a path parameter $s\in[0,1]$ that is formalized via a position spline $p_r(s) \in \mathbb{R}^3$ and an orientation curve $R_r(s)\in SO(3)$.
For the robot joint configuration $q\in\mathbb{R}^n$, the end-effector pose $p_{ee}(q)$ and $R_{ee}(q)$, RMPCC refers to a surrogate path–progress state $(s,v_s)$ with discrete dynamics. For the state and control input
$ x_k = \begin{bmatrix} q_k^\top & s_k & v_{s,k} \end{bmatrix}^\top,  
u_k = \begin{bmatrix} \dot q_k^\top & \dot v_{s,k} \end{bmatrix}^\top,
$
the controller minimizes the tracking error and control effort over a horizon $N$ via
$
J = L_f(x_N) + \sum_{k=1}^{N-1} L(x_k,u_k,u_{k-1})
$
with a stage cost that consists of a term for the tracking, the orientation and the regularization. 


\noindent The error $e = p_r(s_k)-p_{ee}(q_k)$ is split into a contouring and a lag error using the local path tangent 
$\Delta \dot q_k = \dot q_k-\dot q_{k-1}$ (with $\Delta \dot q_1 = 0$). 
The risk is considered via CBFs on (i) kinematic singularities, (ii) self–collision, and (iii) environment collision. 
The algorithm proposes barrier functions $h(q)\ge 0$ for each case as elaborated in \cite{Yoon2025}. Since GUARD addresses environmental collisions, the focus is laid on the corresponding CBF given by
$h_{\text{sing}}(q) = \mu(q)-\varepsilon_{\text{sing}}$; 
and 
\begin{equation}
h_{\text{env},\ell}(q,p_{\text{obs}})=d_{\text{env},\ell}(q,p_{\text{obs}})-r_{\text{obs}}-\varepsilon_{\text{env}}
\end{equation}
for each link $\ell$. Here, $d_{\text{env},\ell}$ is the min. distances to the $\ell$-th robot link's collision with an obstacle as predicted by a deep neural network. The safety margin is denoted by $\varepsilon_{\cdot} > 0$. The optimization is solved via QP relaxation of the barrier inequalities via Jacobian linearization. The Gauss–Newton Hessian approximation enables real-time at 100 Hz. Since RMPCC relies on a \emph{perfectly accurate} perception, GUARD addresses this limitation by adding PKO-ICP. \\

\begin{figure}[t]
\centering
\input{Fig/safeMotionsRiskPipeline/pipeline}
\vspace{0.01cm}
\caption{The pipeline of SafeMotionsRisk. The controller makes use of a task policy trained to complete a specific task and a backup policy trained to avoid collisions. If a defined threshold on the collision risk is exceeded, the controller switches to the backup policy to generate a collision-free trajectory. Adjusted from~\cite{Kiemel2024SafeMotionsRisk}.}
\label{fig:rmpcc}
\end{figure}

\noindent \textbf{SafeMotionsRisk} \\
The SafeMotionsRisk controller by Kiemel et al.~\cite{Kiemel2024SafeMotionsRisk} employs a task policy $\pi^T$ and a backup policy $\pi^B$, both trained via reinforcement learning. The task policy $\pi^T$ is trained to perform a desired robot task (e.g., reaching target points), whereas the backup policy $\pi^B$ is trained to generate collision-free trajectories. In state $s$, the executed action $a$ depends on the estimated collision risk $\bar{C}(s, a^{T})$, where $a^T$ denotes the action produced by the task policy~$\pi^T$.
If the estimated collision risk is below a specified risk threshold $\tau$, action $a^T$ is executed. Otherwise, the backup policy $\pi^B$ is used to generate a backup action $a^B$:

\begin{equation} \label{eq:SafeMotionsRisk}
    a =
    \begin{cases}
      a^{T} & \text{if $\bar{C}(s, a^{T}) \le \tau$} \\
      a^{B}        &\text{otherwise}
    \end{cases}
 \end{equation}


\begin{figure*}[t!]
    \centering
    \includegraphics[scale=0.45]{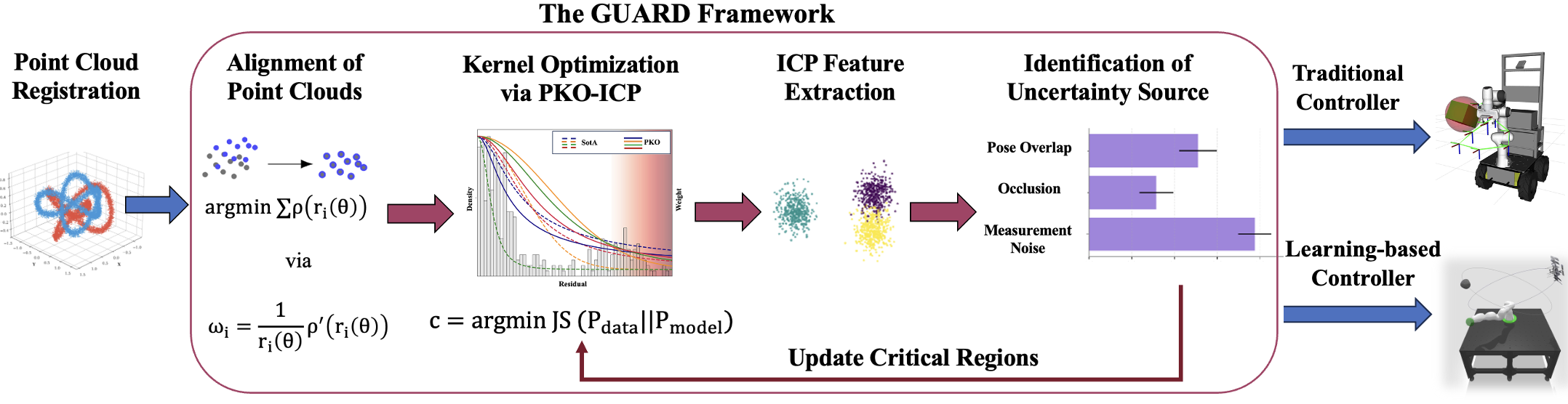}
	\caption{Schematic description of GUARD. The point cloud is aligned via the PKO in \cite{Choi2025}. The output of PKO combined with TCAV-GPC is used to extract relevant features and perform the sensitivity-aware uncertainty attribution method. Based on the estimated uncertainties, the collision margin of a traditional controller and the selected policy of a hybrid learning-based controller are adapted in real-time.} 
	\label{fig:Fig3}
\end{figure*}

\section{The Framework \textit{GUARD}}\label{sec:method}
\noindent This section presents GUARD(\textbf{G}uided robot control via \textbf{U}ncertainty attribution and prob\textbf{A}bilistic kernel optimization for \textbf{R}isk-aware \textbf{D}ecision making). 
Fig. \ref{fig:Fig3} shows the pipeline. 
\subsection{Point Cloud Registration and Alignment }
\noindent The perception is conducted on the basis of RGB-D/LiDAR frames. We assume that the point cloud registration and the ICP are performed prior to applying GUARD. While the experiments are performed on synthetic datasets, we note that the algorithm can be applied to real-world data in a straightforward manner as performed in the works of Gaus et al. in \cite{gaus2025humaninterpretableuncertaintyexplanationspoint} and Choi et al. in \cite{Choi2025}.

\subsection{Uncertainty-aware PKO via Concept Feedback} \label{subsec:PKO_GPC}
\noindent GUARD refers to the PKO in \cite{Choi2025} that assigns weights to the residuals to reduce the influence of large residuals from outliers. Especially, these weights yield enhanced robustness. 
To enable efficient uncertainty estimation, GUARD complements the PKO algorithm with the TCAV-GPC method in \cite{gaus2025humaninterpretableuncertaintyexplanationspoint}. As explained in Section \ref{sec:preliminaries}, the uncertainty sources are obtained via the concept activation functions introduced in \cite{Kim2018}. 
Specifically, the sources are classified into pose overlap, occlusion, or sensor measurement noise. Combining PKO with TCAV--GPC, GUARD optionally extends the tuner to a vector $\boldsymbol\theta_k\in\Theta$ to account for correspondence gate, robust-loss scale, or length-scales additionally to the kernel scale. These additional parameters follow small regularized likelihood steps over a short window within safe bounds. The kernel scale continues with the JS divergence update in Eq. \eqref{eq:PKO}. Hence, the PKO parameters are adapted online via the uncertainty and its source. GUARD initializes the parameter adaptation only if necessary. Once the uncertainty exceeds a specified threshold, small adaptations are applied in the direction that mitigates the cause. Formally, TCAV--GPC returns concept posteriors with the ICP pose $T_k\!\in\!SE(3)$, the residuals $\{r_i\}$, and the vector of registration parameters $\boldsymbol{\theta}_k\!\in\!\Theta$ via 
\begin{equation}
    \boldsymbol{\pi}_k=\{\pi_k(c)\}_{c\in\mathcal C} 
\end{equation}
over $\mathcal C=\{\text{noise},\,\text{pose-error},\,\text{overlap}\}$ and an accumulated uncertainty $u_k$.
GUARD adapts the parameters once a dominant concept and/or non-trivial uncertainty occurs:  
\begin{equation}
\max_c \pi_k(c)\ge\tau_{\text{on}};\;\; \;\;\;\; u_k\ge u_{\min}.    
\end{equation}
A small on/off buffer with $\tau_{\text{off}}<\tau_{\text{on}}$ and a minimum interval between updates prevents rapid toggling. 
If the condition is not met, parameters are not modified such that $\boldsymbol{\theta}_{k+1}=\boldsymbol{\theta}_k$. GUARD refers to small steps along concept-specific directions $\{\mathbf U_c\}$:
\begin{equation}
    \Delta\boldsymbol{\theta}_k \;=\; \eta_k\,s(u_k)\!\sum_{c\in\mathcal C}\! \alpha_c\,\pi_k(c)\,\mathbf U_c,
\end{equation}
where $\eta_k$ is a step size, $s(u_k)$ scales the step with the accumulated uncertainty, and $\alpha_c$ are per-concept gains. 
GUARD selects the exact schedules empirically. 
The updated parameters are projected onto the feasible set $\Theta$:
\begin{equation}
\boldsymbol{\theta}_{k+1} \;=\; \Pi_{\Theta}\!\big(\boldsymbol{\theta}_k+\Delta\boldsymbol{\theta}_k\big).
\end{equation}
\noindent If the control loop is close to its deadline, GUARD skips the adaptation and falls back to the uncertainty computation.
In case an uncertainty source stays dominant for several frames, GUARD varies the viewpoint. The computation $O(|\mathcal C|\,d)$ is negligible compared to the ICP and feature extraction.
\subsection{Risk-aware Robot Control} \label{subsec:RMPCC_ICP}
\noindent The output of GUARD is forwarded to the control frameworks in \cite{Yoon2025} and \cite{Kiemel2024SafeMotionsRisk}. The uncertainty and its source serve as guides for the adaptation of the robot's trajectory for collision avoidance. Particularly, accounting for the uncertainties relaxes the assumption of a perfectly accurate perception. \\ 
\noindent \textbf{Uncertainty-aware CBF Adaptation}:
For the traditional RMPCC, the scalar risk and the per-concept posteriors modulate (i) the CBF relaxation (large slack penalties at high risk), (ii) collision margins and desired speed $v_{\mathrm{des}}$, and (iii) soft weights $w_c,w_l,w_o, w_{v_s}$ to trade efficiency for conservatism when the uncertainty spikes. This preserves RMPCC’s real-time performance with responsive safety margins. GUARD provides a risk measure $\rho_k\in[0,1]$ via concept posteriors of TCAV-GPC and the entropy according to
\begin{equation}
r_{\text{obs}}^{\text{eff}} \!=\ r_{\text{obs}} + \kappa_r \rho_k,\qquad
w_{v_s}^{\text{eff}} \!=\ w_{v_s}(1-\kappa_v \rho_k). \label{eq:inflation} 
\end{equation}
We note that the adaptation of the collision margin can be transferred in a generic manner to any framework with CBFs. \\
\textbf{Uncertainty-aware Initialization of the Backup Policy}:
Similar to RMPCC, \mbox{SafeMotionsRisk} assumes a perfectly accurate perception. To account for the uncertainty, we provide GUARD's output at each control step. Particularly, the estimated uncertainty value initiates the switch to the backup policy. In addition, \emph{SafeMotionsRisk} is provided with the uncertainty source.

\section{Validation Experiments} \label{sec:experiments}
\noindent The validation is split in two parts. First, we combine the PKO algorithm with the TCAV--GPC module and evaluate it against two baseline methods.
Second, the uncertainty with the estimated root causes is transferred to two controllers to study how GUARD impacts safety and the task performance.

\subsection{Uncertainty Estimation via PKO}
\begin{figure}[t]
\centering
\includegraphics[scale=0.22]{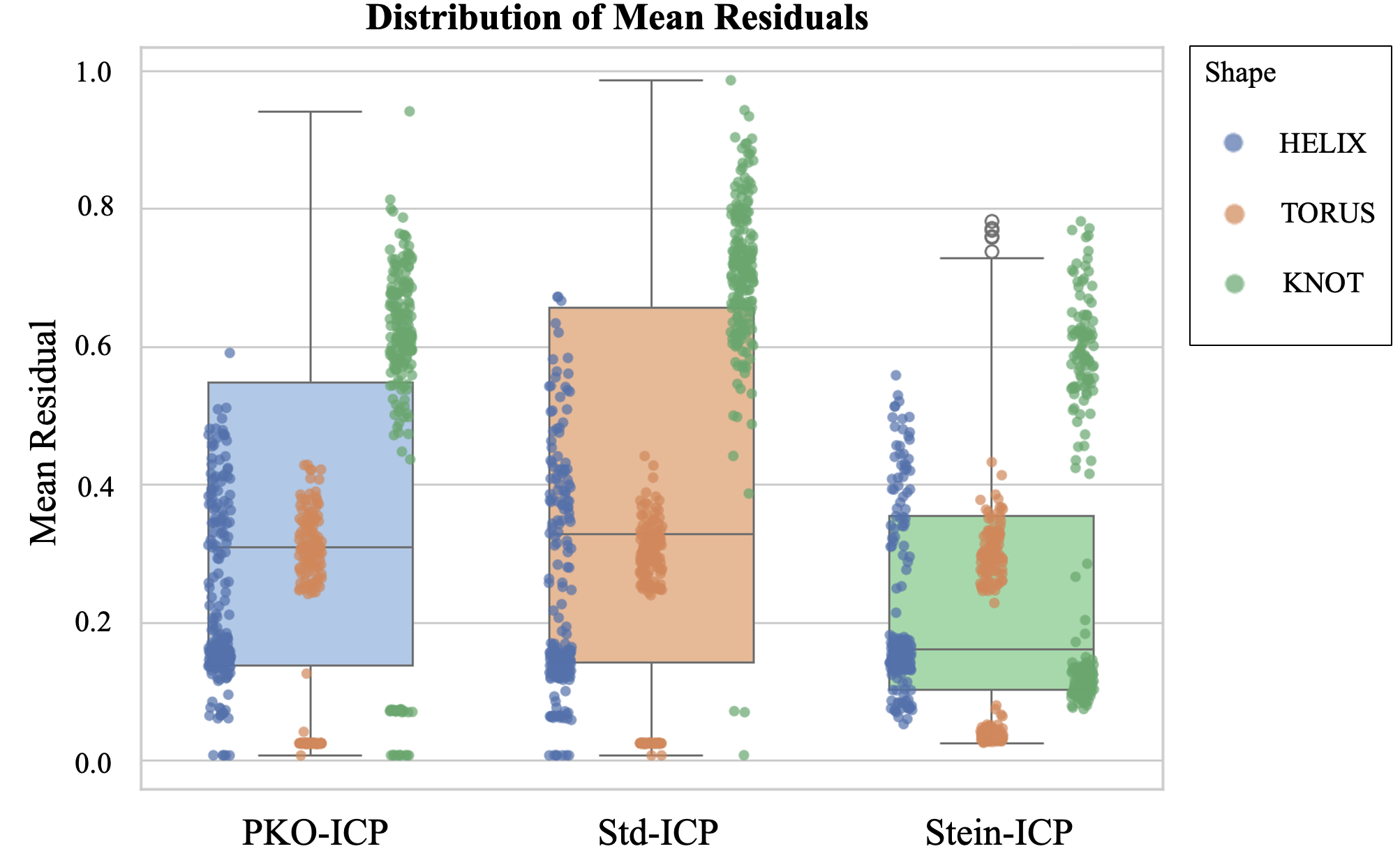}
\caption{Comparison of residuals under injected uncertainty. Distribution of mean residuals for standard ICP, PKO-ICP, and Stein-ICP on synthetic shapes (HELIX, TORUS, KNOT). The PKO-ICP algorithm matches Stein-ICP accuracy trends while remaining faster and more accurate than standard ICP.}
\label{fig:icp_compare}
\end{figure}

\noindent First, we validate the combination of PKO in \cite{Choi2025} and TCAV-GPC in \cite{gaus2025humaninterpretableuncertaintyexplanationspoint}. This loop is referred to by \emph{PKO-ICP}. The experiments evaluate standard point-to-plane ICP \cite{Chen1992}, Stein-ICP \cite{Maken2021}, and PKO-ICP \cite{Choi2025} on a synthetic benchmark of geometric shapes (HELIX, TORUS, KNOT). 
We refer to the residual as the evaluation metric. Ideally, the residuals can be reduced by incorporating their source obtained by TCAV-GPC. 
Figure~\ref{fig:icp_compare} summarizes the mean residuals of the baseline methods and the PKO-ICP. As can be seen, PKO-ICP outperforms standard ICP and closely tracks Stein-ICP in accuracy with tighter interquartile ranges and fewer outliers across all uncertainty classes. 
\noindent Also, one uncertainty source is injected for each run at randomized magnitudes.
\begin{table}[b]
\centering
\normalsize
\setlength{\tabcolsep}{3pt}\renewcommand{\arraystretch}{0.98}
\caption{Latency and mean residuals (approx.).}
\label{tab:latres}
\begin{tabular}{l c c c c}
\hline
\textbf{Method} & \textbf{Time/run} & \textbf{Helix} & \textbf{Torus} & \textbf{Knot} \\
\hline
Std (CPU)   & 0.06--0.15\,s & $\approx 0.25$ & $\approx 0.30$ & $>0.80$ \\
PKO (CPU)   & 0.12--0.35\,s & $\approx 0.20$ & $\approx 0.25$ & $\approx 0.55$ \\
Stein (GPU) & 0.80--1.00\,s & $\approx 0.15$ & $\approx 0.20$ & $\approx 0.30$ \\
\hline
\end{tabular}
\vspace{-2mm}
\end{table}

\noindent\textbf{Runtime:} PKO-ICP achieves a promising trade-off between the accuracy and latency as shown in Tab. \ref{tab:latres}:
Standard ICP offers promising runtimes at higher residuals. 
PKO-ICP achieves $\sim$0.12--0.35\,s with lower residuals, while Stein-ICP is the most accurate but slower.
PKO-ICP is significantly faster than Stein-ICP with comparable residuals and real-time capability.\\
\noindent\textbf{Shape-specific Behavior:} As Fig. \ref{fig:icp_compare} and Tab. \ref{tab:latres} show, PKO reduces the mean residual vs. Standard for the HELIX. Stein ICP yields $\sim$0.05 improvement. TORUS provides the same pattern. For the KNOT, PKO-ICP reduces the residual to $\sim$0.55 and Stein-ICP further to $\sim$0.30.\\
\noindent\textbf{Attribution of Uncertainty Source:} Applying TCAV--GPC on the extracted features in PKO-ICP leads to an assignment of sources to the uncertainties. Following these, PKO focuses on the dominant cause to enhance its robustness.\\
\noindent\textbf{Robustness to Out-of-Distribution (OOD) Data:} GUARD detects OOD frames with a rejection rule on the uncertainty:
\begin{equation}
\text{OOD}_k \;=\; \mathbf{1}\!\Big[\; u_k>\tau_u \;\;\vee\;\; \max_{c\in\mathcal C}\pi_k(c)<\tau_p \;\Big].
\end{equation}
In case of $\text{OOD}_k{=}1$, GUARD freezes kernel updates and sets $\rho_k{\leftarrow}1$. Especially, this reduces $v_{\mathrm{des}}^{\text{eff}}$ while inflating the margins via the respective equations \eqref{eq:inflation}. 



\subsection{Uncertainty-aware Robot Control}

\begin{figure}[t]
  \centering
  \begin{subfigure}{0.3\linewidth}
    \centering
    \includegraphics[width=\linewidth]{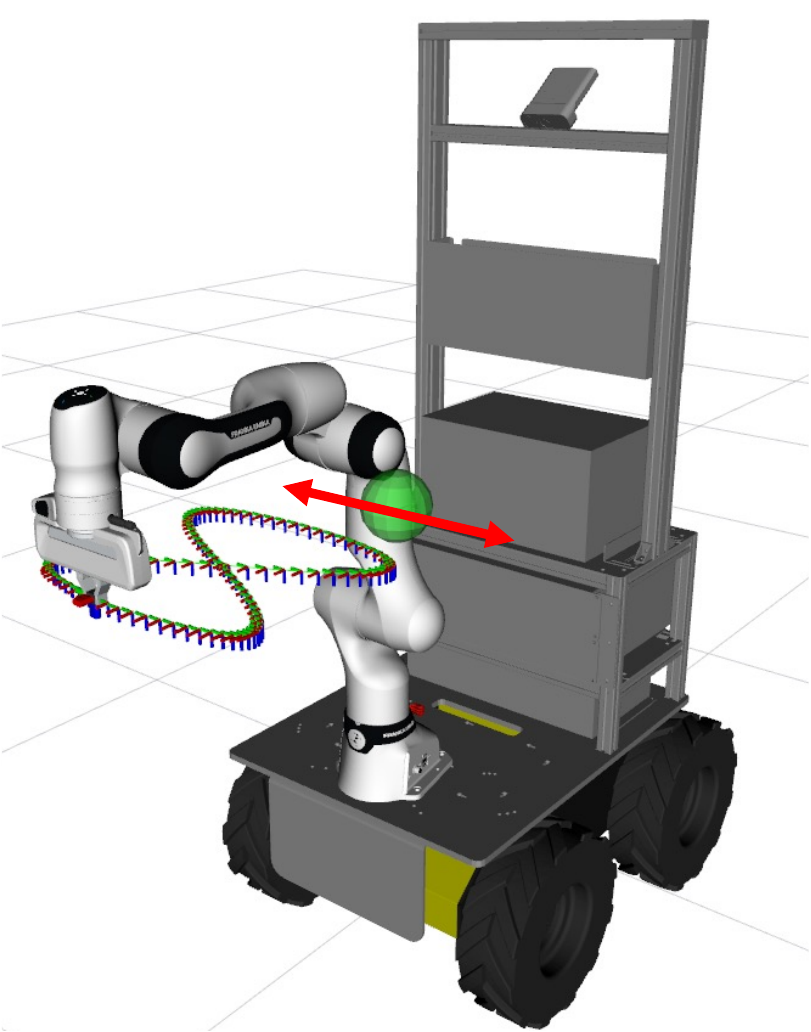}
    \caption{Setup}
    \label{fig:robot_setup}
  \end{subfigure}\hfill
  \begin{subfigure}{0.7\linewidth}
    \centering
    \includegraphics[scale=0.25]{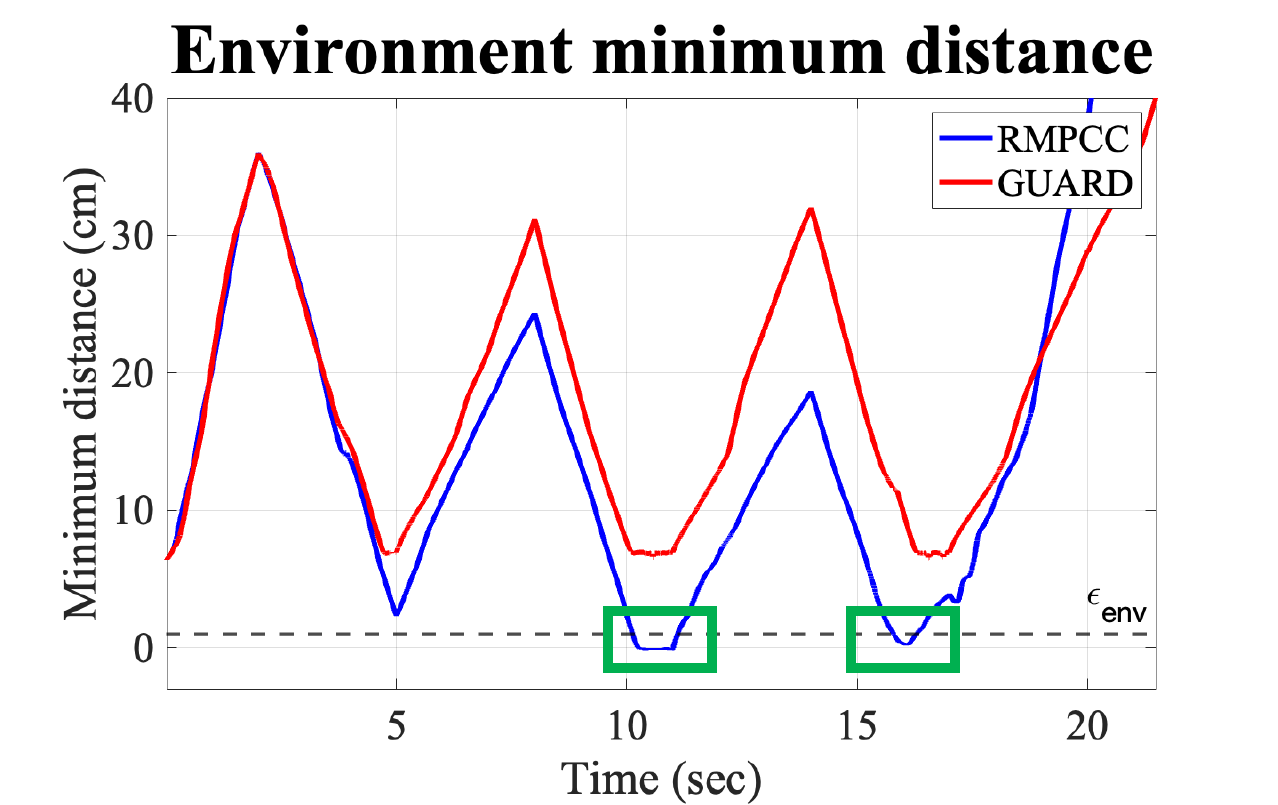}
    \caption{Min. distance}
    \label{fig:env_min_plot}
  \end{subfigure}
  \caption{7-DoF manipulator follows a 3D lemniscate with a dynamic obstacle; (a) setup; (b) minimum robot–obstacle distance for RMPCC (blue) and GUARD (red); dashed: safety margin $\varepsilon_{\text{obs}}$; green boxes indicate collisions.}
  \label{fig:guard_control}
  \vspace{-2mm}
\end{figure}

\setlength{\parindent}{0pt} \paragraph{Validation 1: Traditional RMPCC}
The 7-DoF Panda manipulator Franka Emika was employed, where the robot was provided with a 3D lemniscate path shown in Figure~\ref{fig:robot_setup}. The goal is to avoid a dynamic obstacle. When the obstacle is occluded or not detected, the baseline RMPCC assumes that the obstacle remains stationary at its latest detected position. To replicate this in simulation, we enforced that at randomly chosen time intervals, the robot interpreted the detected obstacle as stationary for a certain duration, while in reality the obstacle continued to move. Here, GUARD adjusts the obstacle’s $r_{\text{obs}}^{\text{eff}}$ and $w_{v_s}$, thereby enabling robust avoidance. Figure~\ref{fig:env_min_plot} illustrates $d_{\text{env}}=\min_{\ell}{d_{\text{env},\ell}}$, the minimum distance between the robot and the obstacle over time. The baseline RMPCC neglects the distance between the detected and actual obstacle positions. Consequently, the min. distance falls below $\varepsilon_{\text{env}}$, resulting in a collision. In contrast, GUARD explicitly considers uncertainties to successfully follow the reference path safely without collisions. To be specific, \noindent GUARD collapses perception into a bounded \emph{risk} $\rho_k\!\in[0,1]$. The controller becomes conservative for high uncertainties. The calibrated concept posteriors $\pi_k(c)$ and entropy $u_k$ yield 
\begin{equation}
\rho_k=\operatorname{clip}_{[0,1]}\!\Big(\beta_0\,u_k+\!\!\sum_{c\in\mathcal C}\!\beta_c\,\pi_k(c)\Big),
\label{eq:risk-simple}
\end{equation}
where $\beta\!\ge\!0$ weight each source. Larger $\rho_k$ linearly inflates margins, increases slack penalty, and reduces speed:
\begin{align}
r_{\mathrm{obs}}^{\mathrm{eff}}&=r_{\mathrm{obs}}+\kappa_r\rho_k, &
\varepsilon_{\mathrm{env}}^{\mathrm{eff}}&=\varepsilon_{\mathrm{env}}+\kappa_\varepsilon\rho_k,
\label{eq:inflation}\\
\gamma^{\mathrm{eff}}&=\gamma_0+\kappa_\gamma\rho_k, &
v_{\mathrm{des}}^{\mathrm{eff}}&=v_{\mathrm{des}}(1-\kappa_v\rho_k).
\label{eq:risk-to-params-simple}
\end{align}
This tightens the safe set and slows motion. 

\setlength{\parindent}{0pt} \paragraph{ Validation 2: GUARD for an RL-based controller}
\label{subsec:safemotions_guard}

Second, we validate GUARD on the hybrid RL controller SafeMotionsRisk~\cite{Kiemel2024SafeMotionsRisk} that switches from a task policy to a backup policy once a probability threshold for collision  is exceeded. We test whether source-aware uncertainty handling reduces collisions and unnecessary activations of the backup policy.

\noindent \textbf{Ablation Studies}\\
At each control step $t$, we inject an artificial uncertainty magnitude $u_t\in[0,1]$ and a synthetic uncertainty source label $\ell_t\in\mathcal{C}$ with $\mathcal{C}=\{\text{sensor noise},\,\text{pose error},\,\text{partial overlap}\}$. We introduce two studies:\\
\textbf {A0 -- Magnitude-based Switching:} The switch to the backup policy is induced if $u_t>\tau$, where $\tau$ denotes the threshold used by SafeMotionsRisk to trigger the backup policy. This baseline treats the uncertainty value as the risk indicator and neglects its source.\\
\textbf{A1 -- Attribution-based Switching and Margin Adaptation}: The controller refers to the predicted uncertainty source $\ell_t$ to adapt its response when $u_t>\tau$:
\begin{itemize}
    \item \textit{Partial overlap}: Increase controller conservatism by enlarging the collision safety margin, making backup activation more likely under sustained overlap-related uncertainty.
    \item \textit{Sensor noise}: Apply temporal smoothing to the uncertainty signal to suppress high-frequency noise and prevent spurious backup activations.
    \item \textit{Pose error}: Immediately activate the backup policy, reflecting a safety-critical failure mode.
\end{itemize}




\noindent \textbf{Experimental Design: Controlled Ablation}\\
To isolate the contribution of source-conditioned switching, we keep the injected uncertainty magnitudes $\{u_t\}$ identical across ablations while varying the labels $\{\ell_t\}$, and measure the effect on the switching behavior. We consider four uncertainty profiles: \emph{Low} ($u_t{=}0.07$), \emph{Medium} ($u_t{=}0.49$), \emph{High} ($u_t{=}0.89$), and \emph{Bursty} ($u_t\in[0.07,0.23]$). 
For each profile and ablation, we run 200 deterministic episodes with fixed seeds and fixed inputs (1600 episodes total, $\sim$56k timesteps). 




\noindent \textbf{Results}\\
We evaluate the goal completion rate, collision rate, and backup activation fraction. Each episode is a reaching task where the robot arm must move its end-effector to a randomly sampled target point within the time of 8s. In case of a collision, the task is terminated. Table~\ref{tab:guard_ablation_safemotions_expanded} summarizes the results.
Importantly, the collision rates show that explicitly accounting for the uncertainty source (A1) leads to notable reductions of the collision rate:
\begin{itemize}
    \item \emph{Low Uncertainty}: A0=$9.5\%$ collision rate vs.\ A1=$3.5\%$ collision rate ($63\% $ reduction). 
    \item \emph{High Uncertainty}: A0 = $10.5\%$ collisions vs.\ A1 = $4.5\%$ collisions ($57\%$ reduction).
    \item \emph{Bursty Uncertainty}: A0 = $9.0\%$ collisions vs.\ A1 = $5.0\%$ collisions ($44\%$ reduction).
\end{itemize}

GUARD yields a reduction of the collision rate by 44--67\% across the evaluated uncertainty profiles compared with the magnitude-only baseline A0. The collision rates of A0 range from 6.0\% to 10.5\%, whereas the attribution-aware approach A1 achieves collision rates between 2.0\% and 5.0\%. The results imply that considering the uncertainty sources reduces collisions by distinguishing cases that can be avoided through adaptation from cases due to fundamentally unsafe regimes that are difficult to avoid in short time windows. 
Providing the uncertainty source enables handling situations with high uncertainties depending on the origin. A1 reduces backup activation when the uncertainty is due to sensor noise because smoothing suppresses short spikes. Also, A1 increases conservatism for partial overlap and reacts immediately in case of pose error.
In accordance with these results, the backup activation decreases in the medium and high profiles (0.067 to 0.033 and 0.133 to 0.033), while collision reductions are maintained. Hence, the uncertainty attribution improves selectivity: The backup policy is less initiated when uncertainty is likely benign (noise) and triggered for safety-critical systematic root causes (pose error, sustained overlap).

\begin{table}[t]
\centering
\setlength{\tabcolsep}{4pt}
\renewcommand{\arraystretch}{0.95}
\caption{SafeMotionsRisk ablation results (200 deterministic episodes per condition). A0: Uncertainty value, A1: uncertainty value and source.}
\label{tab:guard_ablation_safemotions_expanded}
\begin{tabular}{l c c c c}
\hline
\textbf{Profile} & \textbf{Abl.} & \textbf{Goal (\%)} & \textbf{Collision (\%)} & \textbf{Backup Frac.} \\
\hline
Low    & A0 & 96.0 & 9.5 & 0.033 \\
       & A1 & 92.5 & 3.5 & 0.033 \\
\hline
Medium & A0 & 49.0 & 6.0 & 0.067 \\
       & A1 & 51.0 & 2.0 & 0.033 \\
\hline
High   & A0 & 12.0 & 10.5 & 0.133 \\
       & A1 & 9.0  & 4.5  & 0.033 \\
\hline
Bursty & A0 & 79.5 & 9.0  & 0.020 \\
       & A1 & 81.0 & 5.0  & 0.020 \\
\hline
\end{tabular}
\vspace{-2mm}
\end{table}

\noindent \textbf{Interpretation and Key Insights}\\
From the above results, we conclude that accounting for the uncertainty source yields 
\begin{itemize}
    \item \textit{Improved Risk-awareness:} The effective reduction of collisions across profiles indicates that providing the uncertainty sources can guide the backup switch more effectively than magnitude-only thresholding.
    \item  \textit{Enhancement of the Efficiency:} The backup activation is reduced by 50\%--75\% while the goal completion remains comparable. 
\end{itemize}

\subsection{Limitations}
\label{sec:limitations}
\noindent \textbf{Approximation Error and Barrier Shaping}: The linearization of the Jacobian and the Gauss–Newton Hessian may degrade once sharp variations of the kinematics or barriers occur. These effects will be addressed in the future.\\ 
\textbf{PKO-TCAV}: The evaluation is limited in the diversity of scenarios and sensors. The sensitivity of the adaptation law to initialization, hyperparameters, and possible drifts will be studied in the future across kernel priors and update rates.\\
\noindent \textbf{Ablation Studies with SafeMotionsRisk}:
Section \ref{subsec:safemotions_guard} validates the policy switches for settings A1 vs. A0. The protocol refers to fixed seeds for reproducibility. Thus, effects due to stochastic perturbations have been neglected. Furthermore, we referred to synthetic uncertainty profiles in the experiments. The evaluation is limited to reaching tasks in simulation. Future work will address randomized scenario testing in real-world settings and the generalizability to additional robotic subdomains (e.g., navigation, assembly,...). 

\section{Conclusion}
\label{sec:conclusion}
\noindent This paper presented GUARD that combines a kernel optimization algorithm from robot state estimation with an uncertainty attribution technique from the perception domain for safe robot control frameworks. Especially, GUARD provides not only the amount of the uncertainties, but also their sources. In the experiments, we provided GUARD's output to two controllers to derive safety-critical thresholds. In the MPC simulation, GUARD successfully avoided collisions under the evaluated uncertainty conditions. In the RL-based SafeMotionsRisk experiments in \cite{Kiemel2024SafeMotionsRisk}, GUARD reduced collision rates by 44--67\% compared with magnitude-only uncertainty handling. Furthermore, backup policy activations were reduced by approximately 50--75\% in the medium- and high-uncertainty profiles, demonstrating improved efficiency through sensitivity-aware uncertainty attribution. Future work will focus on evaluating GUARD's performance in dynamic real-world settings. 

\bibliographystyle{IEEEtran}
\bibliography{ref}

\end{document}